\documentclass[runningheads,a4paper]{llncs}
\usepackage{graphicx}
\usepackage{amsmath,amssymb} 
\usepackage{color}

\usepackage{epsfig}
\usepackage{float}
\usepackage{multirow}

\DeclareMathOperator*{\argmin}{arg\,min}
\DeclareMathOperator*{\argmax}{arg\,max}

\begin{document}
\mainmatter


\title{How important are ``Deformable Parts'' \\ in the Deformable Parts Model?} 

\author{Santosh K. Divvala, Alexei A. Efros, Martial Hebert}
\institute{Carnegie Mellon University}

\maketitle

\begin{abstract}


The main stated contribution of the Deformable Parts Model (DPM) detector of Felzenszwalb et al. (over the Histogram-of-Oriented-Gradients approach of Dalal and Triggs) is the use of deformable parts. A secondary contribution is the latent discriminative learning. Tertiary is the use of multiple components. A common belief in the vision community (including ours, before this study) is that their ordering of contributions reflects the performance of detector in practice. However, what we have experimentally found is that the ordering of importance might actually be the reverse. First, we show that by increasing the number of components, and switching the initialization step from their aspect-ratio, left-right flipping heuristics to appearance-based clustering, considerable improvement in performance is obtained. But more intriguingly, we show that with these new components, the part deformations can now be completely switched off, yet obtaining results that are almost on par with the original DPM detector. Finally, we also show initial results for using multiple components on a different problem -- scene classification, suggesting that this idea might have wider applications in addition to object detection.
\end{abstract}

\section{Introduction}
\label{sec:intro}


\begin{figure}
\centering
\setlength{\tabcolsep}{0.1pt}
\begin{tabular}{cc}
\multicolumn{2}{c}{\includegraphics[height=1.2in]{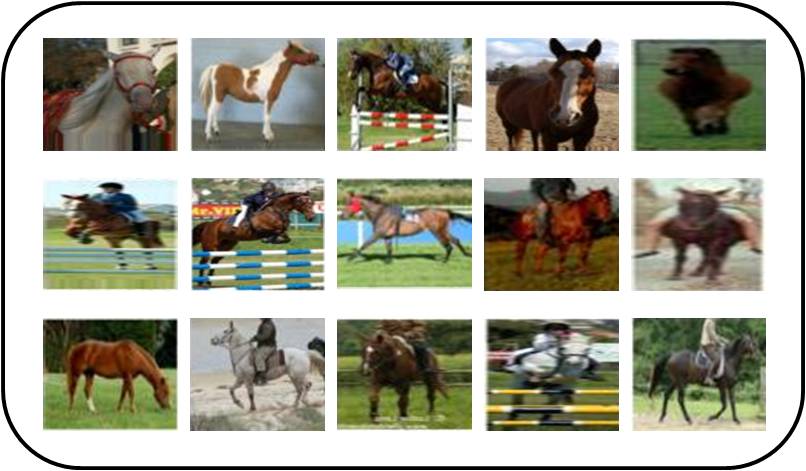}}  \\
\multicolumn{2}{c}{{\footnotesize Monolithic Classifier~\cite{dalal2005}}} \\
\includegraphics[height=0.83in]{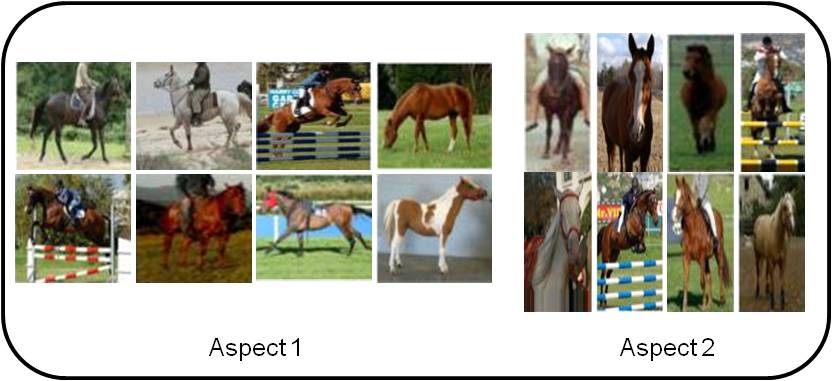} &
\includegraphics[height=0.95in]{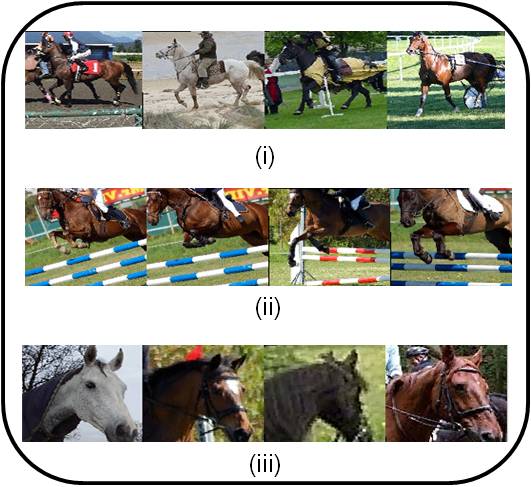} \\
{\footnotesize Aspect-ratio split~\cite{Felzenszwalb10,Park10}} & {\footnotesize Poselet split~\cite{Bourdev10}} \\ 
\includegraphics[height=0.83in]{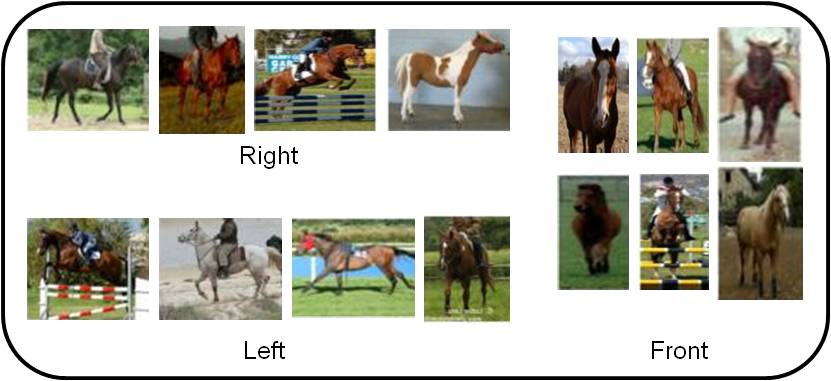} &
\includegraphics[height=1.03in]{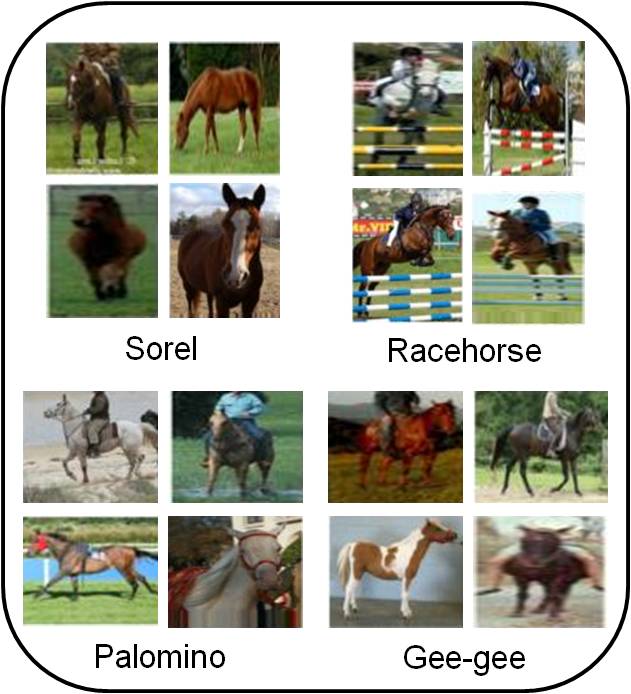} \\
{\footnotesize Viewpoint split~\cite{Chum07a,Ren10}} & {\footnotesize Taxonomy split~\cite{ImageNet,Ferrari11}} \\
\multicolumn{2}{c}{\includegraphics[height=1.8in]{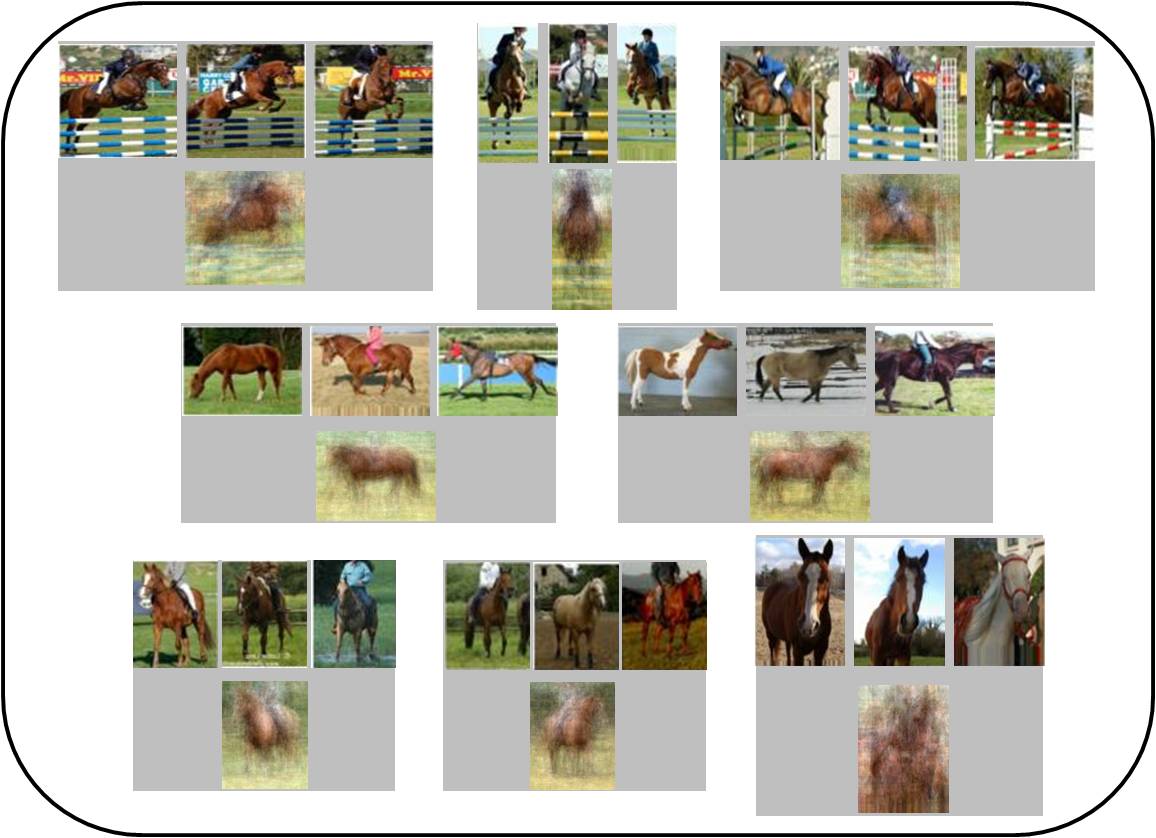}}  \\
 \multicolumn{2}{c} {{\footnotesize Visual Subcategories (this paper)}} \\
\end{tabular}
\caption{The standard {\em monolithic classifier} is trained on all instances together. {\em Viewpoint split} partitions the training data using ground-truth viewpoint annotations into left, right, and frontal subcategories. {\em Poselets} clusters the instances based on ground-truth keypoint annotations in the configuration space.  {\em Taxonomy split} groups instances into subordinate categories using a human-defined semantic taxonomy. {\em Aspect-ratio split} uses a very simple bounding box aspect-ratio heuristic. {\em Visual subcategories} are obtained using (unsupervised) appearance-based clustering (top: few examples, bottom: mean image)}
\label{fig:horseteaser}
\end{figure}

Consider the images of category horse in Figure~\ref{fig:horseteaser} (row1) from the challenging PASCAL VOC dataset~\cite{Pascal_VOC}. Notice the huge variation in the appearance, shape, pose and camera viewpoint of the different horse instances -- there are left and right-facing horses, horses jumping over a fence in different directions, horses carrying people in different orientations, close-up shots, etc. How can we build a high-performing sliding-window detector that can accommodate the rich diversity amongst the horse instances?

Deformable Parts Models (DPM) have recently emerged as a useful and popular tool for tackling this challenge. The recent success of the DPM detector of Felzenszwalb et al.,~\cite{Felzenszwalb10} has drawn attention from the entire vision community towards this tool, and subsequently it has become an integral component of many classification, segmentation, person layout and action recognition tasks (thus receiving the lifetime achievement award at the PASCAL VOC challenge).

Why does the DPM detector~\cite{Felzenszwalb10} perform so well? As the name implies, the main stated contribution of~\cite{Felzenszwalb10} over the HOG detector described in~\cite{dalal2005} is the idea of deformable parts. Their secondary contribution is latent discriminative learning. Tertiary is the idea of multiple components (subcategories). The idea behind deformable parts is to represent an object model using a lower-resolution `root' template, and a set of spatially flexible high-resolution `part' templates. Each part captures local appearance properties of an object, and the deformations are characterized by links connecting them. Latent discriminative learning involves an iterative procedure that alternates the parameter estimation step between the known variables (e.g., bounding box location of instances) and the unknown i.e., {\em latent} variables (e.g., object part locations, instance-component membership). Finally, the idea of subcategories is to segregate object instances into disjoint groups each with a simple (possibly semantically interpretable) theme e.g., frontal vs profile view, or sitting vs standard person, etc, and then learning a separate model per group. 

A common belief in the vision community is that the deformable parts is the most critical contribution, then latent discriminative learning, and then subcategories. Although the ordering somewhat reflects the technical novelty  (interestingness) of the corresponding tools and the algorithms involved, is that really the order of importance affecting the performance of the algorithm in practice?
 

What we have experimentally found from our analysis of the DPM detector is that the ordering might actually be the reverse! First, we show that (i) by increasing the number of subcategories in the mixture model, and (ii) switching from their aspect-ratio, left-right flipping heuristics to appearance-based clustering, considerable improvement in performance is obtained. But more intriguingly, we show that with these new subcategories,  the  part deformations can be completely turned off, with only minimal performance loss. These observations together highlight that the conceptually simple subcategories idea is indeed an equally important contribution in the DPM detector that can potentially alleviate the need for deformable parts for many practical applications and object classes.


\section{Understanding Subcategories}
\label{sec:review}

In order to deal with significant appearance variations that cannot be tackled by the deformable parts, \cite{Felzenszwalb10} introduced the notion of multiple components i.e., subcategories into their detector. The first version of their detector~\cite{Felzenszwalb08} only had a single subcategory. The next version~\cite{Felzenszwalb10} had two subcategories that were obtained by splitting the object instances based on aspect ratio heuristic. In the latest version~\cite{Felzenszwalb_voc-release4}, this number was increased to three, with each subcategory comprising of two bilaterally asymmetric i.e., left-right flipped models (effectively resulting in 6 subcategories). The introduction of each additional subcategory has resulted in significant performance gains (e.g., see slide 23 in~\cite{Felzenszwalb11talk}). 

Given this observation, what happens if we further increase the number of subcategories in their model? In Section~\ref{sec:objdetres}, we will see that this does not translate to improvement in performance. This is because the aspect-ratio heuristic does not generalize well to a large number of subcategories,  and thus fails to provide a good initialization. Nonetheless, it is possible to explore other ways to generate subcategories. For example, subcategories for cars can be based either on object pose (e.g., "left-facing", "right-facing", "frontal"), or car manufacturer (e.g., Subaru, Ford, Toyota), or some functional attribute (e.g., sports car, utility vehicle, limousine). Figure~\ref{fig:horseteaser} illustrates a few popular subcategorization schemes for horses. 

\begin{figure}
\centering
\begin{tabular}{cc}
\includegraphics[width=0.2\linewidth]{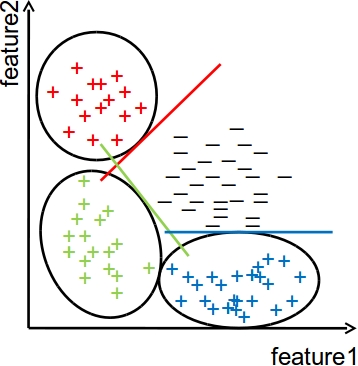} &
\includegraphics[width=0.2\linewidth]{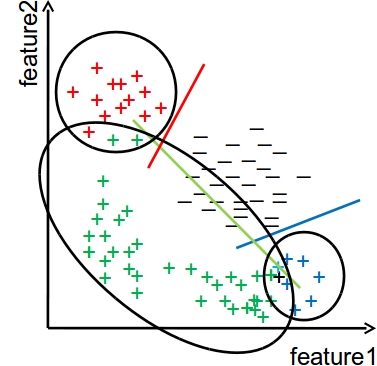} \\
{\footnotesize Visual Subcategories} \ \ \ & \ \ \ {\footnotesize Semantic Subcategories} \\
\end{tabular}
\caption{{\footnotesize A single linear model cannot separate the data well into two classes. (Left) When similar instances (nearby in the feature space) are clustered into subcategories, good models can be learned per subcategory, which when combined together separate the two classes well. (Right) In contrast, a semantic clustering scheme also partitions the data but leads to subcategories that are not optimal for learning the category-level classifier.}}
\label{fig:visvssem}
\end{figure}

What is it that the different partitioning schemes are trying to achieve? A closer look at the figures reveals that they are trying to encode the homogeneity in appearance. It is the {\em visual homogeneity} of instances within each subcategory that simplifies the learning problem leading to better-performing classifiers (Figure.~\ref{fig:visvssem}). What this suggests is, instead of using semantics or empirical heuristics, one could directly use appearance-based clustering for generating the subcategories. We use this insight to define new subcategories in the DPM detector, and refer to them as {\em visual} subcategories (in contrast to semantic subcategories that involve either human annotations or object-specific heuristics). \\


\noindent{\bf Related Work} The idea of subcategories is inspired by works in machine learning literature~\cite{Jacobs91,Chen08,Japkowicz02,Xu05,Fu08,Fradkin08} that consider solving a complex (nonlinear) classification problem by using locally linear classification techniques. Several computer vision approaches have explored different strategies for generating subcategories. In~\cite{SchneidermanK00,Chum07a, Harzallah08}, viewpoint annotations associated with instances were used to segregate them into separate left, right, frontal sub-classes. In~\cite{Park10}, the size (height) of detection windows was used to cluster them into near and far-scale sub-classes. In~\cite{Yang11}, co-watch features are used to group videos of a specific category into simpler subcategories. In~\cite{Bourdev10}, instances are clustered into {\em poselets} using keypoint annotations in the configuration space. In~\cite{ImageNet}, subordinate categories of a basic-level category are constructed using human annotations. 

The concept of subcategories has also received significant attention in cognitive psychology~\cite{Johnson98,RoLl78}. In the seminal work of~\cite{RoLl78}, the idea of prototypes was introduced. The prototype concept relies on the notion of typicality: its resemblance to the other members of the category and its differences to the members of other categories. 
 
Closely related are also the recently popular exemplar-based methods~\cite{Chum07a,Malisiewicz11}. While in a global strategy, a single classifier is trained using all instances belonging to a class as positives, in the case of exemplar-based methods, a separate classifier is learned for each individual instance. Although promising results have been demonstrated, exemplar methods are prone to overfitting since too much emphasis is often placed on local irregularities in the data~\cite{Vilalta04}. The global and local learning strategies sit at two extremes of a large spectrum of possible compromises that exploit information from labeled examples. This paper explores intermediate points of this spectrum.

\section{Learning Subcategories}
\label{sec:subcat}

We first briefly review the key details of using subcategories in the DPM detector, and then explain the details specific to their use in our analysis.


Given a set of $n$ labeled instances (e.g., object bounding boxes) $D = (<x_1,y_1>, \ldots, <x_n,y_n>)$, with $y_i \in \{-1, 1\}$, the goal is to learn a set of $K$ subcategory classifiers to separate the positive instances from the negative instances, wherein each individual classifier is trained on different subsets of the training data. The assignment of instances to subcategories is modeled as a latent variable $z$. This binary classification task is formulated as the following (latent SVM) optimization problem that minimizes the trade-off between the $l_2$ regularization term and the hinge loss on the training data~\cite{Felzenszwalb10}:
\begin{eqnarray} \label{lsvm1}
 \small
 \argmin_w \frac{1}{2} \sum_{k=1}^{K} ||w_k||^2 + C \sum_{i=1}^n \epsilon_i, \\
  y_i . s_i^{z_i} \geqslant 1 - \epsilon_i, \ \epsilon_i \geqslant 0, \\
  z_i = \argmax_k s_i^k, \\
  s_i^k = w_k.\phi_k(x_i)+b_k.
\end{eqnarray}
The parameter $C$ controls the relative weight of the hinge-loss term, $w_k$ denotes the separating hyperplane for the kth subclass, and $\phi_k(.)$ indicates the corresponding feature representation. Since the minimization is semi-convex, the model parameters $w_k$ and the latent variable $z$ are learned using an iterative approach~\cite{Felzenszwalb10}. \\


\noindent{\bf Initialization} As mentioned earlier, a key step for the success of latent subcategory approach is to generate a good initialization of the subcategories. Our initialization method is to warp all the positive instances to a common feature space $\phi(.)$, and to perform unsupervised clustering in that space. In our experiments, we found the Kmeans clustering algorithm using Euclidean distance function to provide a good initialization. \\

\begin{figure}
\centering
\begin{tabular}{cc}
\includegraphics[width=0.49\linewidth]{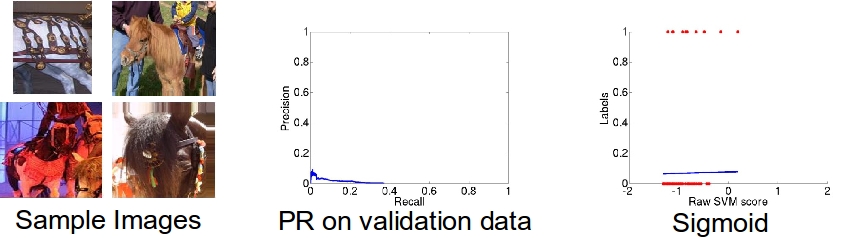} &
\includegraphics[width=0.49\linewidth]{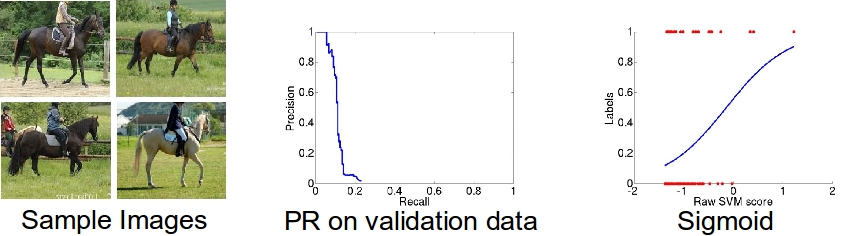} \\
(a) `Noisy' Subcategory & (b) `Good' Subcategory
\end{tabular}
\caption{{\footnotesize The classifier trained on a noisy subcategory (horses with extreme occlusion and confusing texture) performs poorly on the validation dataset. As a result, its influence is suppressed by the sigmoid. While a good subcategory (horses with homogeneous appearance) classifier leads to good performance on the validation data and hence its influence is boosted by the calibration step.}}
\label{fig:calibsig}
\end{figure}

\noindent{\bf Calibration} One difficulty in merging subcategory classifiers during the testing phase is to ensure that the scores output by individual SVM classifiers (learned with different data distributions) are calibrated appropriately, so as to suppress the influence of noisy ones. We address this problem by transforming the output of each SVM classifier by a sigmoid to yield comparable score distributions~\cite{platt2000,Wang11}\footnote{Even though the bias term $b_k$ is used in equation (\ref{lsvm1}) to make the scores of multiple classifiers comparable, we have found that it is possible for some of the subcategories to be very noisy (specifically when $K$ is large), in which case their output scores cannot be compared directly with other, more reliable ones.}(Figure~\ref{fig:calibsig}). Given a thresholded output score $s_i^k$ for instance $i$ in subcategory $k$, its calibrated score is defined as 
\begin{equation} \label{calib0}
g_i^k=\frac{1}{1+exp(A_k.s_i^k + B_k)},
\end{equation}
where $A_k,B_k$ are the learned parameters of the following logistic loss function:
\begin{eqnarray} \label{calib1}
 \small
\argmin_{A_k,B_k} \sum_{i=1}^n t_i \log g_i^k + (1-t_i) \log(1-g_i^k), \\
t_i = Or(W_i^k, W_i).
\end{eqnarray}
$Or(w_1,w_2)=\frac{|w_1 \cap w_2|}{|w_1 \cup w_2|} \in [0,1]$ indicates the overlap score between two bounding boxes~\cite{Ferrari11b}, $W_i$ is the ground-truth bounding box for the ith training sample, and $W_i^k$ indicates the predicted bounding box by the kth subcategory. In our experiments, we found this calibration step to help improve the performance (mean A.P. increase of 0.5\% in the detection experiments).

\section{Experimental Analysis}
\label{sec:objdetres}

\begin{table}[t]
\begin{tiny}
\begin{center}
\begin{tabular}{|l||c|c|c|c|c|c|c|c|c|c|c|c|c|c|c|c|c|c|c|c|c|}
\hline
 & aero & bike & bird & boat & botl & bus & car & cat & chr & cow & table & dog & horse & mbik & pers & plant & sheep & sofa & trn & tv & {\em MEAN} \\ \hline 
 ~\cite{Felzenszwalb_voc-release4} & 28.9 & 59.5 & 10.0 & 15.2 & 25.5 & 49.6 & 57.9 & 19.3 & 22.4 & 25.2 & 23.3 & 11.1 & 56.8 & 48.7 & 41.9 & 12.2 & 17.8 & 33.6 & 45.1 & 41.6 & 32.3 \\ \hline
 ours & 32.7 & 58.5 & 6.6 & 17.5 & 28.5 & 51.7 & 56.4 & 26.7 & 24.4 & 27.3 & 33.2 & 13.0 & 58.5 & 50.0 & 41.2 & 15.4 & 24.5 & 32.7 & 49.7 & 43.3 & 34.6 \\ \hline
 ~\cite{Felzenszwalb_voc-release4}-p & 25.9 & 46.8 & 9.7 & 13.8 & 18.6 & 42.8 & 39.8 & 13.0 & 17.1 & 21.1 & 16.4 & 10.3 & 55.3 & 42.6 & 36.2 & 11.5 & 16.3 & 27.7 & 42.4 & 38.2 & 27.3 \\ \hline
 ours-p & 27.8 & 54.3 & 10.8 & 15.2 & 24.0 & 48.4 & 50.0 & 19.8 & 18.2 & 22.9 & 29.0 & 11.8 & 55.4 & 44.8 & 37.3 & 14.7 & 21.6 & 27.0 & 44.6 & 41.3 & 30.9 \\ \hline 
\end{tabular}
\end{center}
\caption{{\scriptsize Results on VOC2007. Row1: result of~\cite{Felzenszwalb_voc-release4}. row2: result using visual subcategories. row3: result of~\cite{Felzenszwalb_voc-release4} with parts turned off i.e., using all the features at twice the spatial resolution of the root filter with no deformations. row4: same result using visual subcategories.}}
\label{tab:subcatres}
\end{tiny}
\end{table}

\begin{figure}
\centering
\begin{tabular}{c} 
\includegraphics[width=0.9\linewidth]{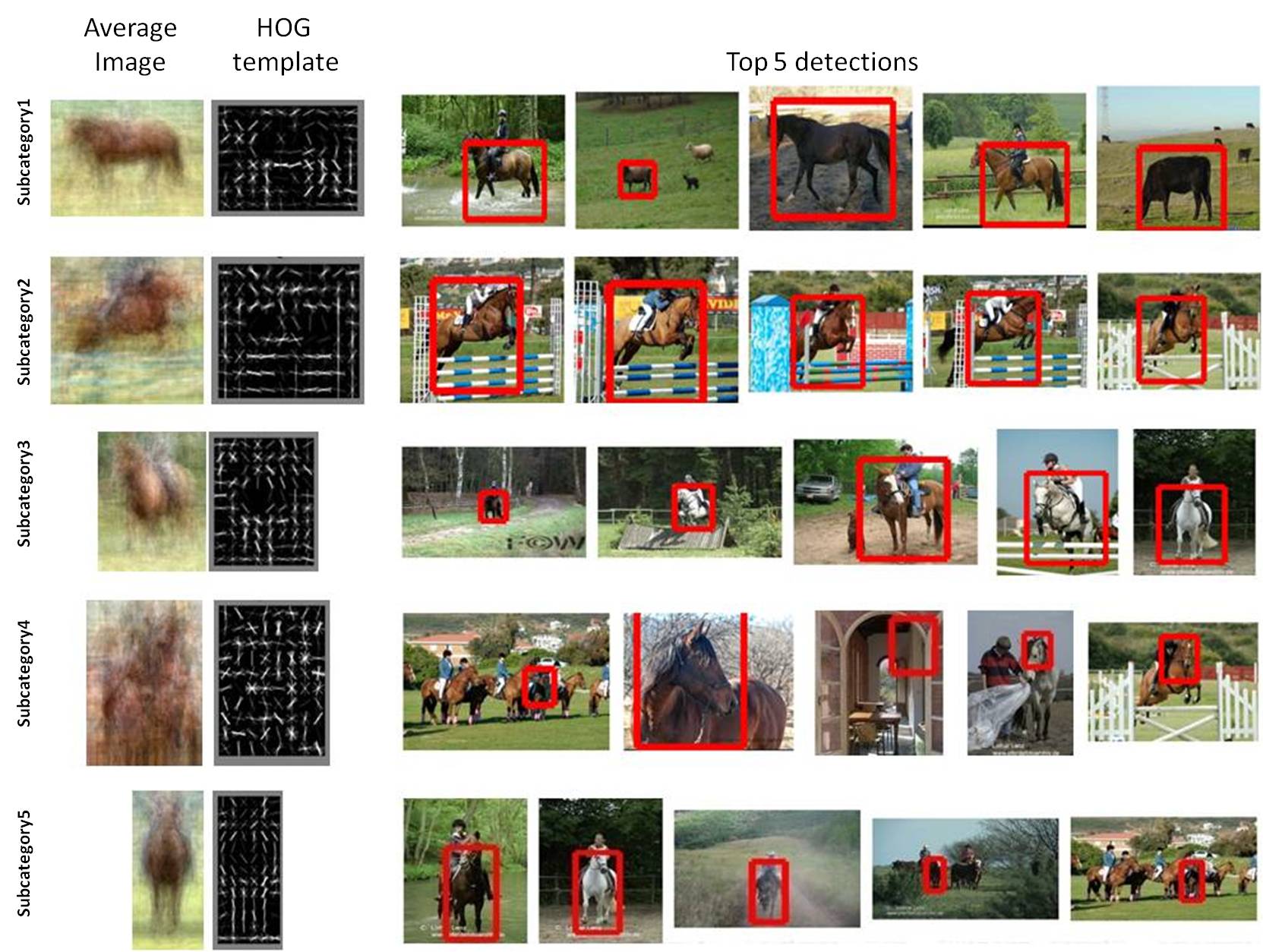}  \\
{Category: Horse} \\
\includegraphics[width=0.9\linewidth]{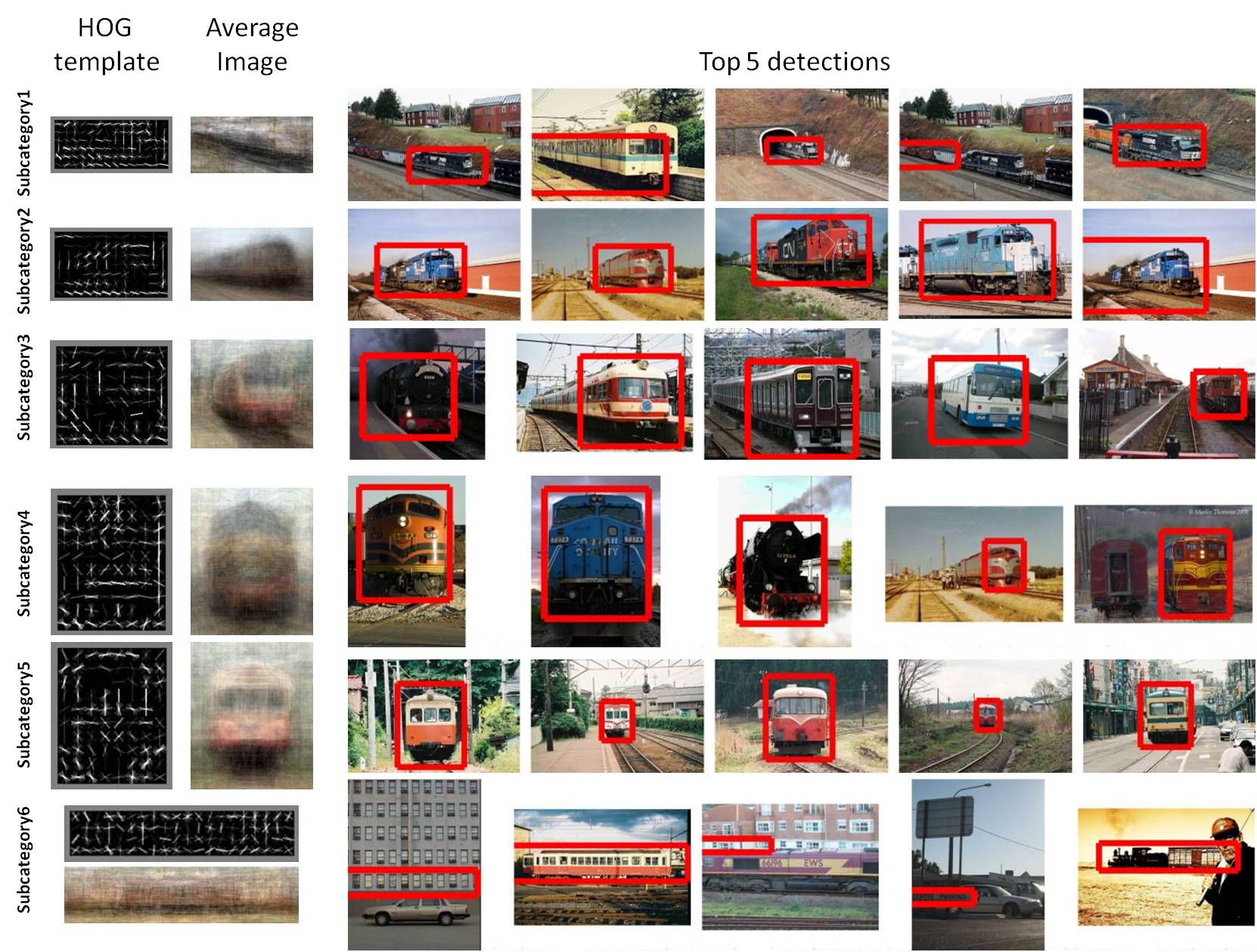}  \\ 
 {Category: Train}
\end{tabular}
\caption{{\footnotesize As the intra-class variance within subcategories is low, the learned detectors perform quite well at localizing instances of their respective subcategories. Notice that for the same aspect-ratio and viewpoint, there are two different subcategories (rows 4,5) discovered for the train category.}}
\label{fig:percompresults}
\end{figure}

\begin{figure}
 \centering
 \begin{tabular}{c}
 \includegraphics[width=0.66\linewidth]{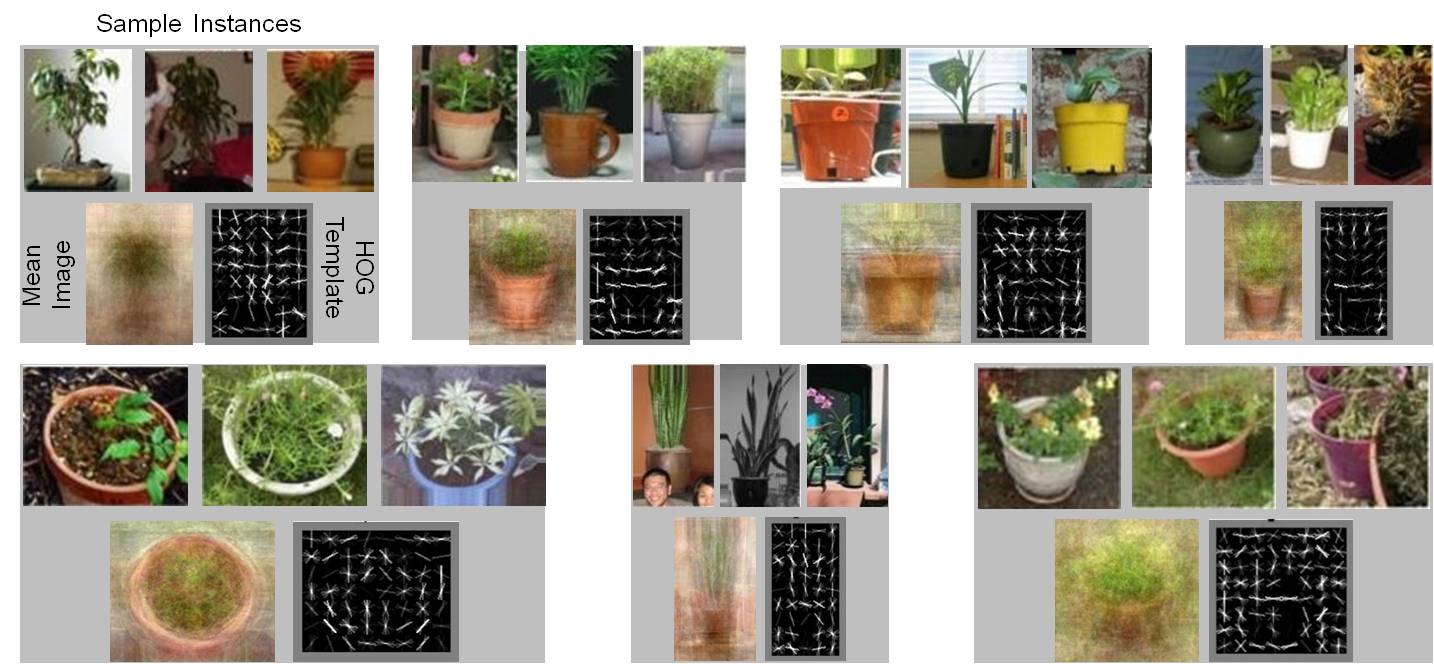} \\ 
 \includegraphics[width=0.66\linewidth]{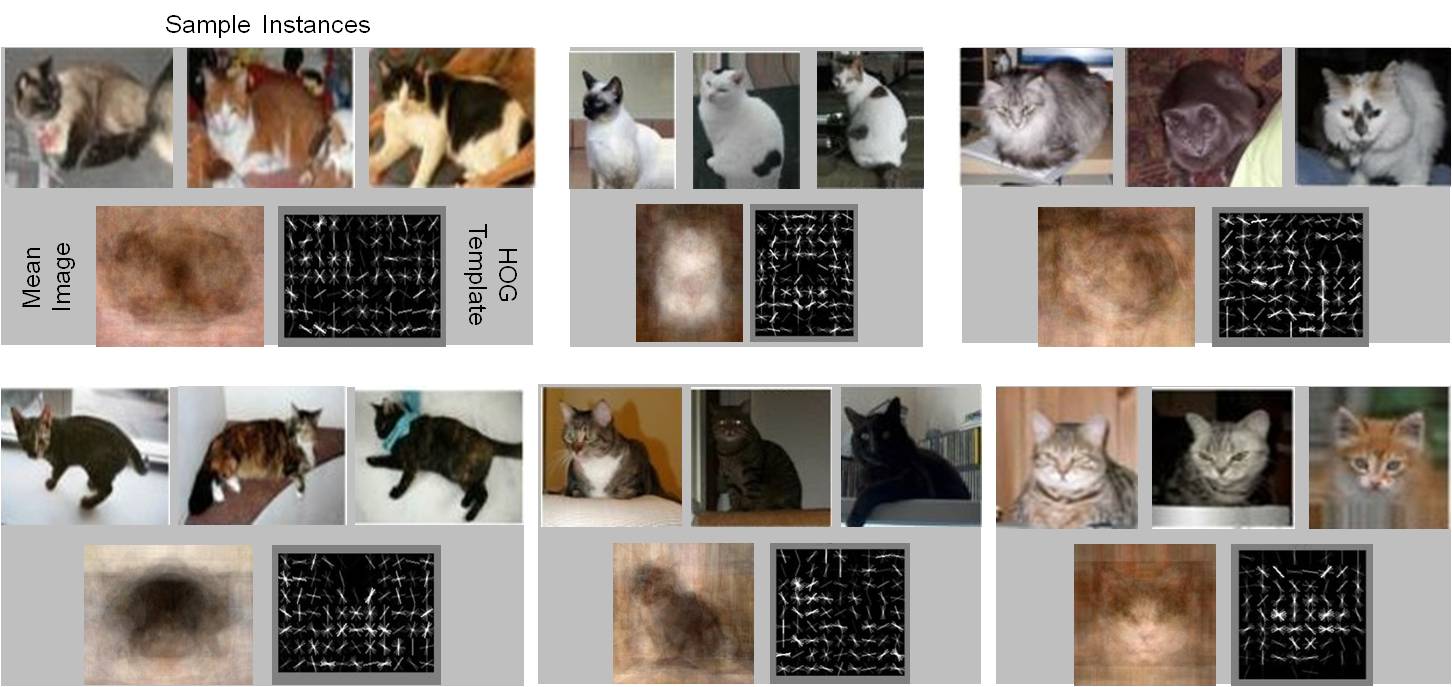}   
 \end{tabular}
 \caption{{\scriptsize The visual subcategories discovered for pottedplants correspond to different camera viewpoints, while cats are partitioned based on their pose. The baseline system~\cite{Felzenszwalb_voc-release4} based on the aspect-ratio, left-right flipping heuristic cannot capture such distinctions (as many of the subcategories share the same aspect-ratio and are symmetric).}}
 \label{fig:clusters}
 \end{figure}

We performed our analysis on the PASCAL VOC 2007 comp3 challenge dataset and protocol~\cite{Pascal_VOC}. Table~\ref{tab:subcatres} summarizes our key results. Row1 shows the (baseline) result of the DPM detector~\cite{Felzenszwalb_voc-release4}. Row2 shows the result obtained by using visual subcategories (with $K$=15) in the DPM detector. It surpasses the baseline by 2.3\% on average across the 20 VOC classes (the mean A.P. improves from 32.3\% to 34.6\%). Figure~\ref{fig:percompresults} shows the top detections obtained per subcategory for horse and train categories. The individual detectors do a good job at localizing instances of their respective subcategories. In Figure~\ref{fig:clusters}, the discovered subcategories for symmetric (pottedplant) and deformable (cat) classes are displayed. The subcategories obtained for all of the 20 VOC classes are displayed in the supplementary material.

Rows 3,4 of table~\ref{tab:subcatres} show the results obtained by turning off the deformable parts. More specifically, rather than sampling `parts' from the high-resolution HOG template (sampled at twice the spatial resolution relative to the features captured by the root template) and modeling the deformation amongst them, we directly use all the features from the high-resolution template. This update to the DPM detector results in a simple multi-scale (two-level pyramid) representation with the finer resolution catering towards improved feature localization. We observe that using this two-level pyramid representation for the visual subcategories yields a mean A.P. of 30.9\% that is almost on par as the full deformable parts baseline (32.3\%). This result becomes intiutive from the observation that instances within each of the subcategory are well-aligned (see figure~\ref{fig:clusters} and supplementary material), and thus simpler models (without deformations) would suffice for training discriminative detectors. For instance, in case of rigid objects such as pottedplants, tvmonitors and trains, the use of part deformations does not offer any improvement over using the multi-scale visual subcategory detector. For a few classes though, such as person and sofa, part deformations seem to be useful, while for some others, such as dining-table and sheep, the multi-scale visual subcategories actually performs better. These observations suggest that, in practice, the relatively simple concept of visual subcategories is as important as the use of deformable parts in the DPM detector. \\


\noindent {\bf Computational Issues.} In terms of computational complexity, the two-scale visual subcategory detector ($K$=15) involves one coarse (root) and one fine resolution template per subcategory, totaling a sum of 30 HOG templates. Whereas the DPM detector has $K$=6 subcategories each with one root and eight part templates, totaling 54 HOG templates, which need to be convolved at test time. In terms of model learning, the DPM detector has the subcategory, as well as the part deformation parameters (four) as latent variables (for each of the 48 parts), while the visual subcategory detector only has the subcategory label as latent. Therefore it not only requires fewer rounds of latent training than required by the DPM detector (leading to faster convergence), but also is less susceptible to getting stuck in a bad local minima~\cite{Pawan10}. As emphasized in~\cite{Felzenszwalb10}, simpler models are preferable, as they can perform better in practice than rich models, which often suffer from difficulties in training. \\

\begin{figure}
\centering
\begin{tabular}{cccc}
 \includegraphics[width=0.2\linewidth]{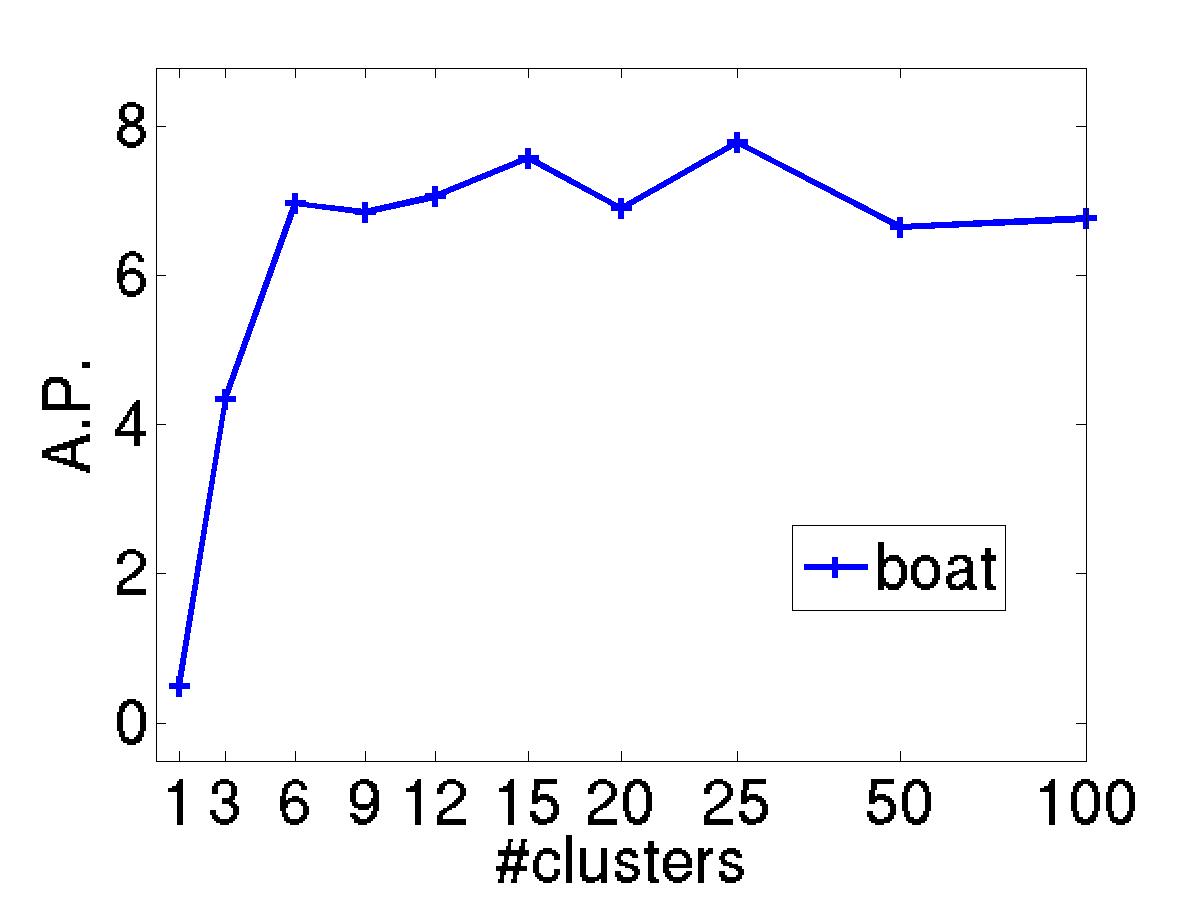} &
 \includegraphics[width=0.2\linewidth]{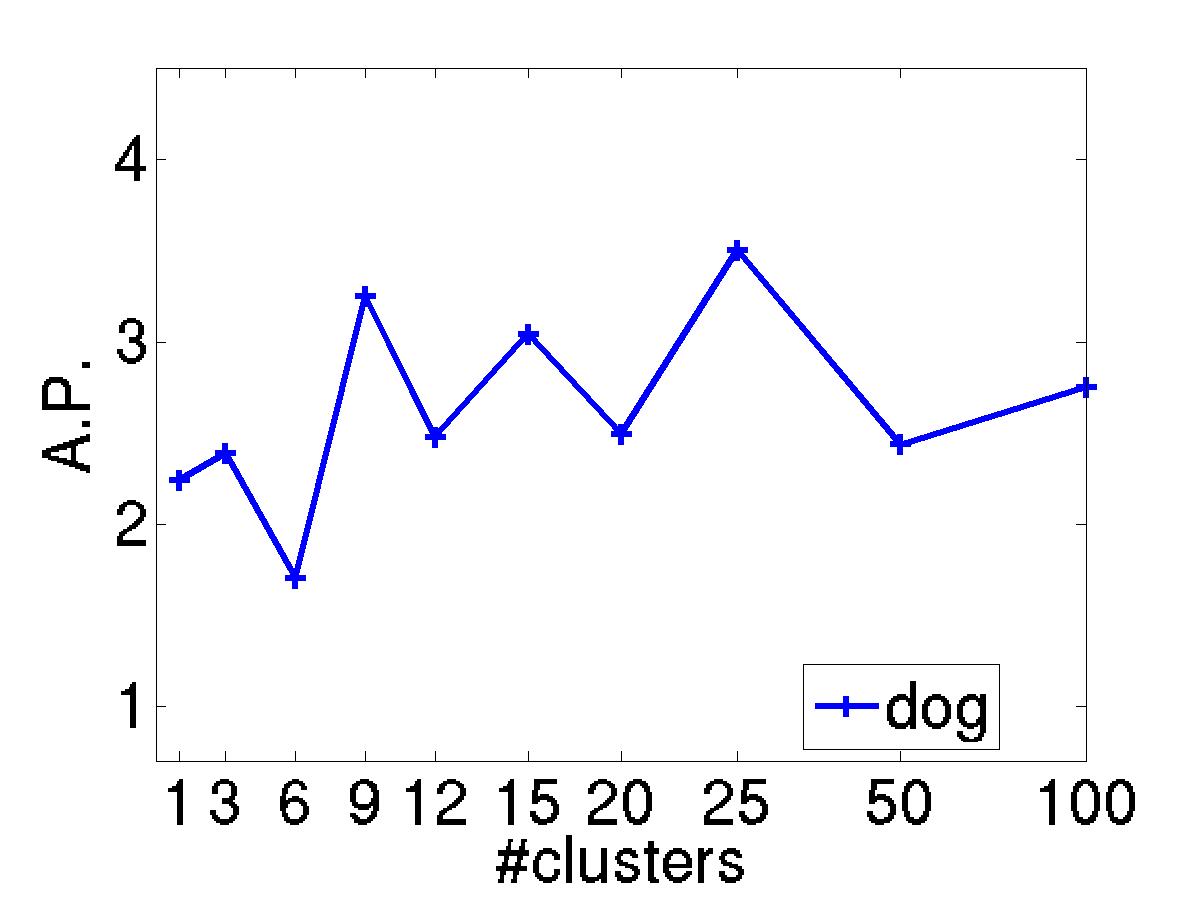}  &
 \includegraphics[width=0.2\linewidth]{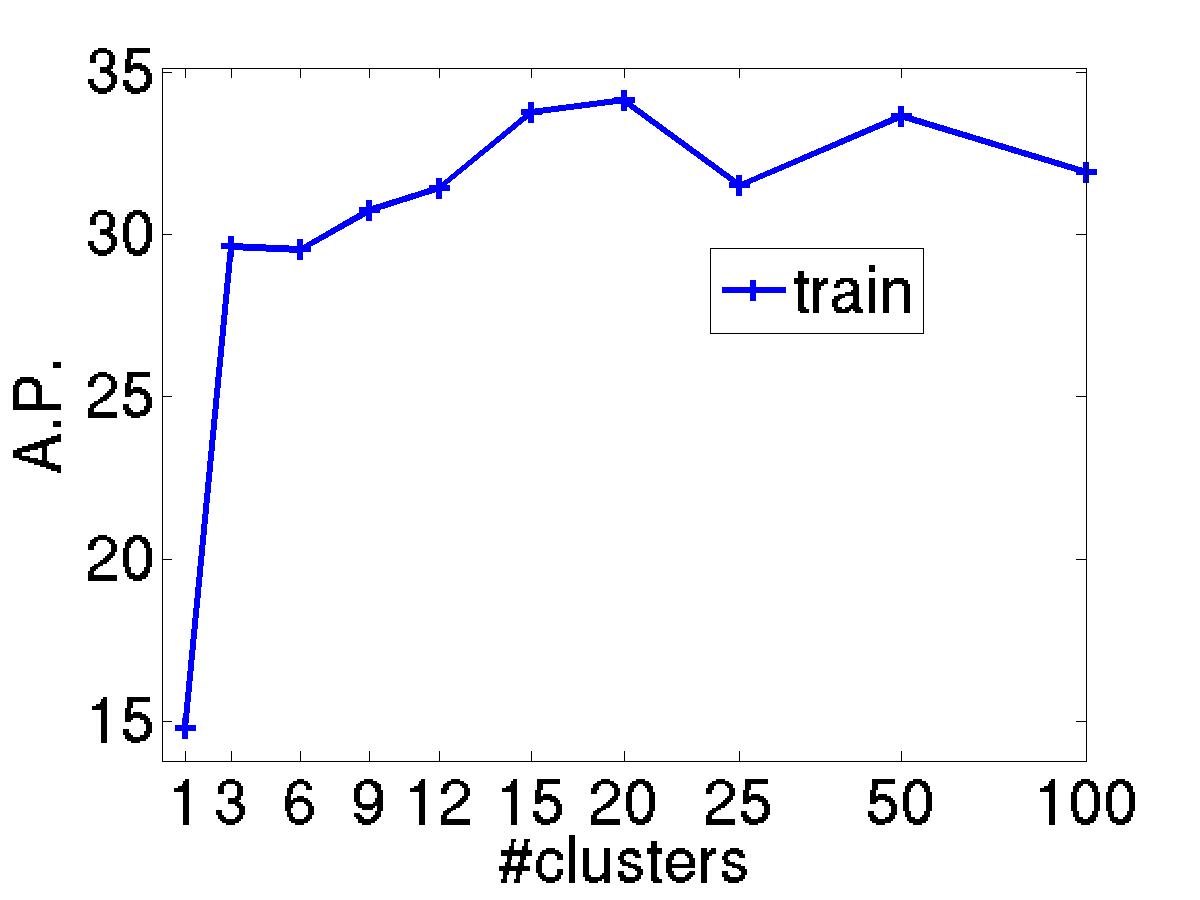} & 
 \includegraphics[width=0.2\linewidth]{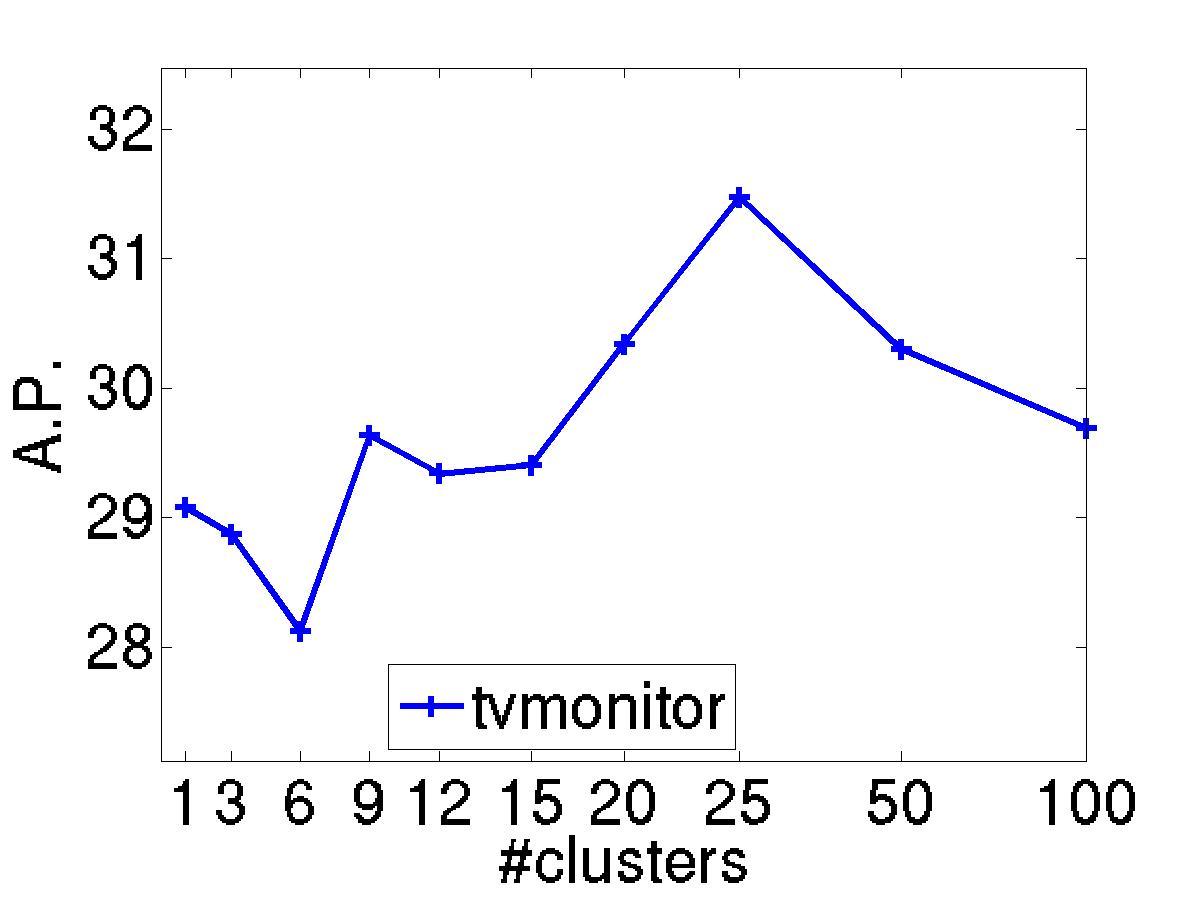}
\end{tabular}
\caption{{\footnotesize  Variation in detection accuracy as a function of number of clusters for four distinct VOC2007 classes. The A.P. gradually increases with increasing number of clusters and stabilizes beyond a point.}}
\label{fig:variation}
\end{figure}

\noindent {\bf Number of subcategories.} One important parameter is the number of subcategories $K$. We analyze the influence of $K$ by using different values ($K=[3,6,9,12,15,20,25,50,100]$) for a few classes (`boat', `dog', `train', `tv') on the validation set. We plot the variation in the performance over different $K$ in figure~\ref{fig:variation}. The performance gradually increases with increasing K, but stabilizes around K=15. We used $K=15$ in all the detection experiments. \\

\noindent {\bf Initialization.} Proper initialization of clusters is a key requirement for the success of latent variable models. We analyzed the importance of appearance-based initialization by comparing it with the aspect-ratio based initialization of~\cite{Felzenszwalb10}. Simply increasing the number of aspect-ratio based clusters leads to a decrease in performance (mean A.P drops by 1.2\%), while for the same number ($K$=15), appearance-based clustering helps improve the mean A.P. by 2.3\%. 

We noticed minimal variation in the final performance on multiple runs with different Kmeans initialization. We found the (latent) discriminative reclustering step helps in cleaning up any {\em mistakes} of the initialization step. (Also we observed that most of the reclustering happens in the first latent update.)


\section{Application: Visual Subcategories for Scene Classification}
\label{sec:scres0}

Another scenario where the problem of high intra-class variability is witnessed is scene classification. Scene categories exhibit a large range of visual diversity due to significant variation in camera viewpoint and scene structure. For example, when we refer to the scene category `coast' (from~\cite{SUN10}), it could contain images of 'rocky' shores, sunsets, cloudy beaches, or calm waters. From our analysis of visual subcategories on the object detection dataset, we could expect that their use could also aid in simplifying the learning task for scene classification. \\

\noindent {\bf Dataset details.} We use the Scene Understanding (SUN) database for our scene classification experiments. The SUN database is a collection of about 100,000 images organized into an exhaustive set of 899 scene categories~\cite{SUN10}. For our experiments, we use the subset of 397 well-sampled categories. These 397 fine-grained scene categories are arranged in a 3-level tree: with 397 leaf nodes (subordinate categories) connected to 15 parent nodes at the second level (basic-level categories) that are in turn connected to 3 nodes at the third level (superordinate categories) with the root node at the top. This hierarchy was not considered in the original experimental evaluations in~\cite{SUN10} but used as a human organizational tool (in order to facilitate the annotation process e.g., annotators navigate through the three-level hierarchy to arrive at a specific scene type (e.g. `bedroom') by making relatively easy choices (e.g. `indoor' versus `outdoor' at the higher level)). 

Our goal is to train a classifier that can identify images as belonging to one of the 15 basic-level categories.\footnote{The 15 basic-level categories are `shoppingNdining', `workplace', `homeNhotel', `vehicleInterior', `sportsNleisure', `cultural', `waterNsnow', `mountainsNdesert', `forestNfield', `transportation', `historicalPlace', `parks', `industrial', `housesNgardens', `commercialMarkets'.} We use the images from all the subordinate categories in a basic-level category to build the data corresponding to that basic-level category. The data was split into half training and half testing. The classifiers are all trained in a `one-vs-all' fashion where instances belonging to a specific category are considered positive examples, and the rest of the instances (belong to all the other categories except the chosen one) serve as negative examples in the training process. While training subcategory classifiers, instances belonging to a particular subcategory (within a category) are treated as positives and the rest of the instances belong to that category are ignored (treated as {\em don't care} examples). The number of subcategories $K$ was set to be 50 (to tackle the larger intra--category diversity in this dataset). We evaluate performance using the A.P. metric as used in PASCAL VOC image classification task~\cite{Pascal_VOC}.


We use the GIST feature representation (using the implementation of~\cite{Oliva01}), that has been well-studied in literature for scene classification experiments (e.g.,~\cite{Ferrari11}). We create this descriptor for each image at a $10 \times 10$ grid resolution where each bin contains that image patch's average response to steerable filters at 8 orientations and 4 scales. We acknowledge that a classification system based on this representation is unlikely to beat the prevailing state-of-the-art. Multiple previous methods have shown that classification performance is significantly improved by the use of BOW-models with densely sampled feature points along with multiple sets of feature descriptors, and the use of spatial pyramids. We chose the simple GIST-based representation for our analysis in this paper as our focus is not to argue in favor of a new classification method, but to show the benefits of the visual subcategory concept using a generic and simple framework.

\subsection{Visual subcategories are semantically interpretable}
\label{sec:scres1}

\begin{figure}
\centering
\begin{tabular}{c|c}
\includegraphics[width=0.5\linewidth]{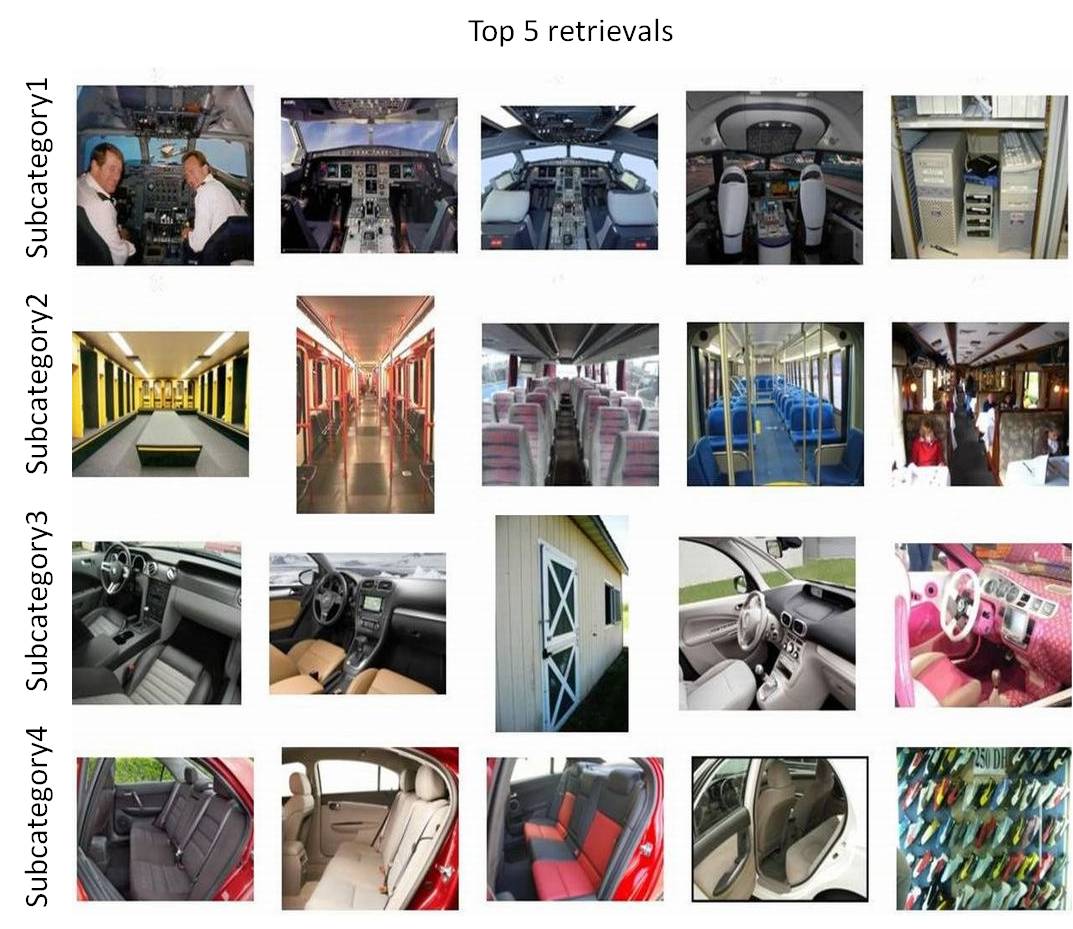}  &  
\includegraphics[width=0.5\linewidth]{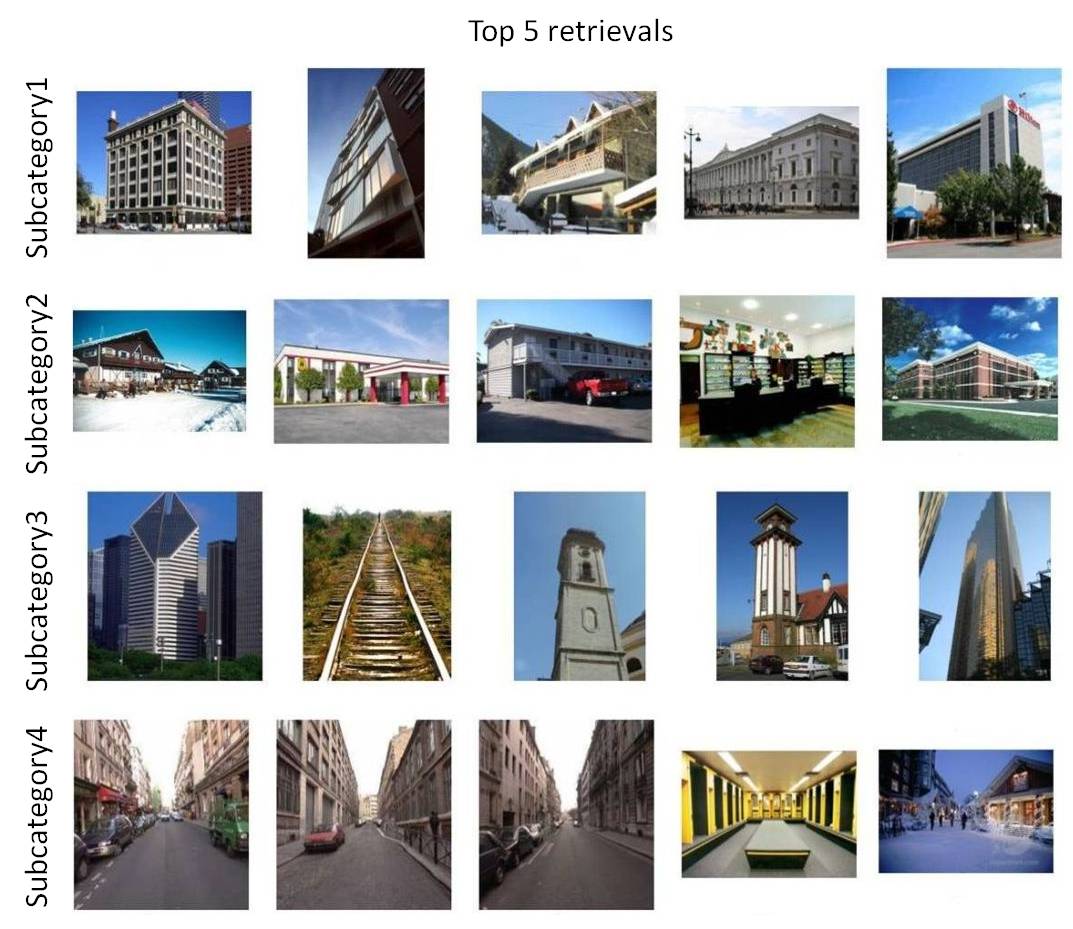} \\
{\footnotesize Category: VehicleInterior} & {\footnotesize Category: CommericalMarket} \\ \hline \\
\includegraphics[width=0.5\linewidth]{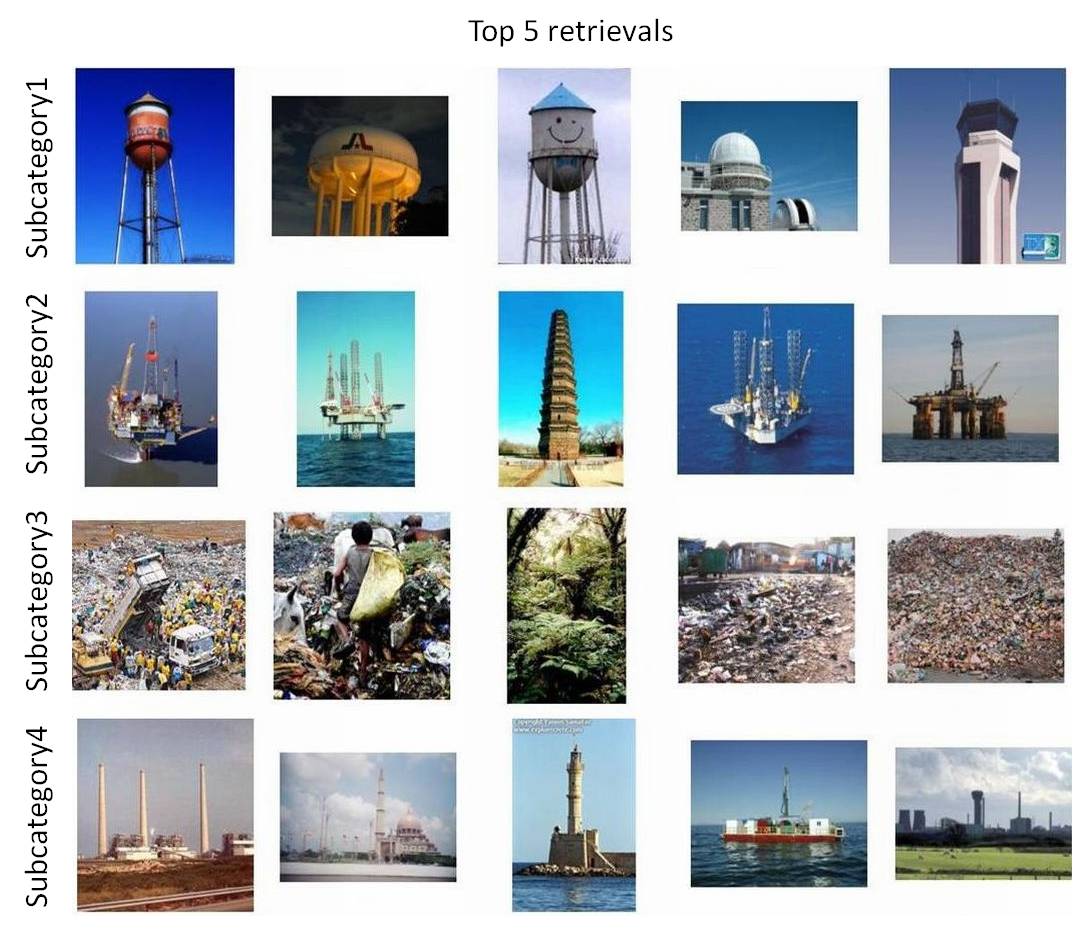}  &  
\includegraphics[width=0.5\linewidth]{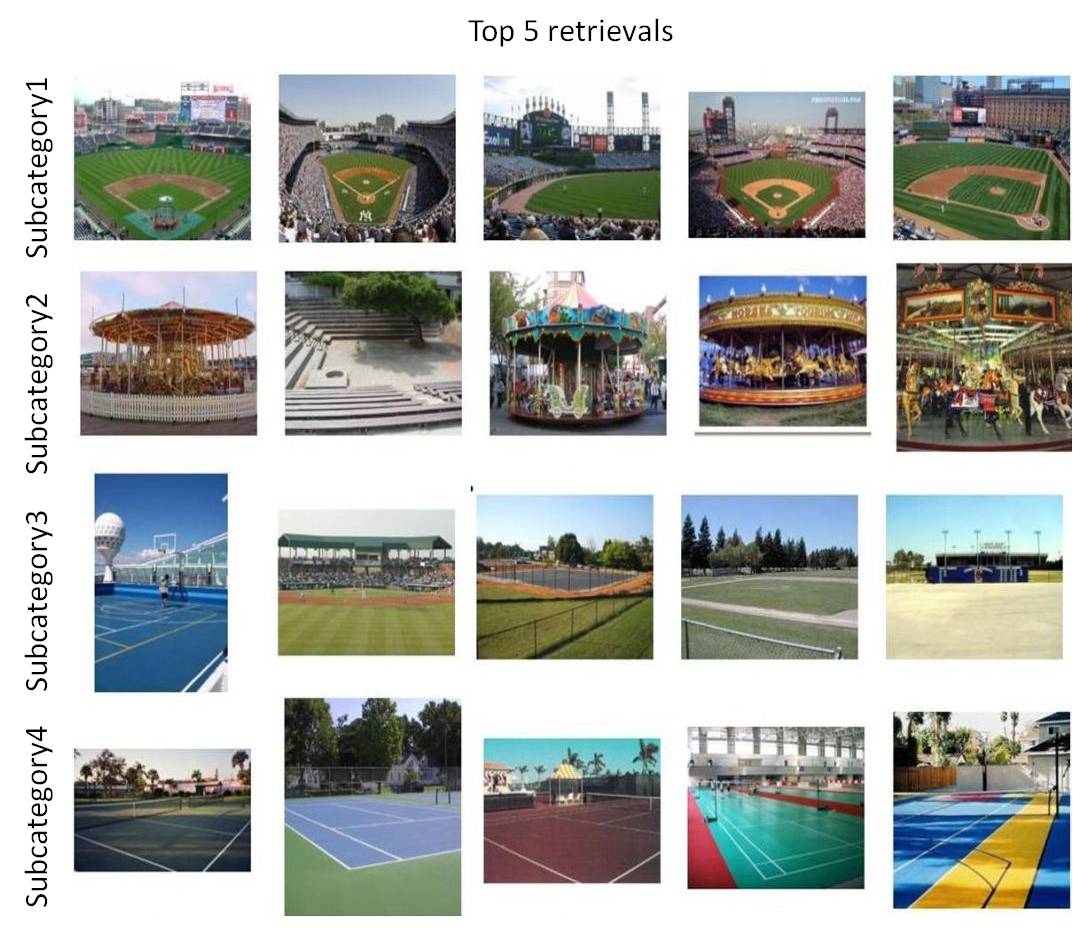} \\
{\footnotesize Category: Industrial} & {\footnotesize Category: Parks} \\
\end{tabular}
\caption{SUN397 Scene-Image Classification Results: Scene-Image categories exhibit a large visual diversity due to significant variation in camera viewpoint and scene structure. The `vehicleInterior' classifier contains separate subcategories for cockpit, bus interior, car front seat and car back seat. `commercialMarket' is composed of different types of buildings, skyscrapers, and street/alley scenes. The `industrial' category has water towers, oil rigs, land fills, and outdoor industrial scenes. Finally the category `park' has baseball fields, carousals, outdoor tennis fields as subcategories. It is interesting to note that using a completely unsupervised approach, it is possible to discover the subcategories that mostly correspond to the human annotated fine-grained categories of the SUN397 dataset (even subtle ones such as car frontseat, car backseat).}
\label{fig:scenes2}
\end{figure}

Our approach based on visual subcategories achieves a score of 27.1\% confidently outperforming the baseline linear SVM 16.9\% (see Table in supplementary material for results per category). The utility of our approach becomes more evident as we take a closer look at the classification results of discovered subcategories (see Figure~\ref{fig:scenes2}). Many of the subcategories discovered correspond to the semantic subordinate categories. For example, the basic-level category `vehicleInterior' contains clusters for `cockpit', `bus interior', `car front seat', and `car back seat' that all correspond to the fine-grained categories constituting this basic-level category. Subsequently, this allows deeper reasoning about the image rather than simply
assigning the category label. For e.g., instead of simply classifying an image as 'vehicleInterior', we could now say that it is a `cockpit' image.


\subsection{Visual subcategories alleviate the need for human supervision}
\label{sec:visvssemsun}

Given the above result, we seek to quantitatively analyze the benefit of gathering human-annotated subordinate categories over the unsupervisedly discovered visual subcategories. To this end, we ran an experiment where the subcategories in our framework are initialized using the ground-truth subordinate categories. The result obtained using this initialization (mean A.P. of 27.2\%) is very similar to that obtained using our unsupervised subcategories of 27.1\% (see Table in supplementary material for results per category). This is interesting because it indicates that human supervision for creating the fine-grained subcategories to train a basic-level category classifier may not be of great benefit compared to the unsupervised visual subcategories. Our observations here are also supported by the recent findings in~\cite{Ferrari11}, wherein semantic similarity was found to be correlated to visual similiarity at the bottom of the ImageNet~\cite{ImageNet} hierarchy i.e., when the basic-level category is sliced into extremely small subsets. However to acquire these fine-grained subcategories, one needs to expend significant amount of human annotation effort.


\section{Conclusion}
\label{sec:concl}

Contrary to the existing belief that deformable parts is the key contribution for the success of the deformable parts model detector, we have found that the use of subcategories can potentially alleviate their need. The use of visual subcategories not only benefits model learning and  performance, but also leads to simpler and more interpretable models. In addition to object detection, their use can also benefit the scene classification task as it can alleviate the need for human supervision in carving the space of fine-grained subordinate categories.

{\scriptsize
\bibliographystyle{splncs}
\bibliography{global}
}

\end{document}